\documentclass[11pt]{article}
\usepackage[a4paper,margin=1in]{geometry} 

\usepackage[utf8]{inputenc}
\usepackage{amsmath}
\usepackage{graphicx}
\usepackage{amssymb} 
\usepackage{url}

\usepackage[above,below]{placeins}  %
\usepackage{randtext}    %

\usepackage[backend=biber,style=authoryear,
maxcitenames=2,
maxbibnames=99,
doi=false,url=false,isbn=false,dashed=false,eprint=false, %
uniquename=false,uniquelist=false,
date=year,        %
backref=false,
sortcites=false   %
]{biblatex}

\DeclareFieldFormat{postnote}{#1}%

\usepackage[raggedright]{titlesec}  %

\title{	
	Communicative need modulates competition in language change
}

\date{\vspace{-30pt}}
\author{Andres Karjus\textsuperscript{1}, Richard A. Blythe\textsuperscript{1,2}, Simon Kirby\textsuperscript{1}, Kenny Smith\textsuperscript{1}\\ \small \textsuperscript{1} Centre for Language Evolution, School of Philosophy, Psychology and Language Sciences,\\ \small University of Edinburgh;\\ \small \textsuperscript{2}School of Physics and Astronomy, University of Edinburgh \\ \footnotesize \tt \randomize{a.karjus}@sms.ed.ac.uk, \{r.a.blythe, simon.kirby, kenny.smith\}@ed.ac.uk } 

\begin{document}

\maketitle

\begin{abstract}
	All living languages change over time. The causes for this are many, one being the emergence and borrowing of new linguistic elements. Competition between the new elements and older ones with a similar semantic or grammatical function may lead to speakers preferring one of them, and leaving the other to go out of use.
	We introduce a general method for quantifying competition between linguistic elements in diachronic corpora which does not require language-specific resources other than a sufficiently large corpus. This approach is readily applicable to a wide range of languages and linguistic subsystems. Here, we apply it to lexical data in five corpora differing in language, type, genre, and time span. We find that changes in communicative need are consistently predictive of lexical competition dynamics. Near-synonymous words are more likely to directly compete if they belong to a topic of conversation whose importance to language users is constant over time, possibly leading to the extinction of one of the competing words. By contrast, in topics which are increasing in importance for language users, near-synonymous words tend not to compete directly and can coexist. This suggests that, in addition to direct competition between words, language change can be driven by competition between topics or semantic subspaces. 
\end{abstract}

\section{Introduction}\label{sec_intro}

The literature on language change is full of examples of new elements, such as borrowings or morphological alternatives, replacing previous variants with similar functions. In English for example, past tense regular forms have been replacing irregular ones and vice versa \parencite[][]{pinker_future_2002}, a number of speech sounds were swapped out with other ones during the the Great Vowel Shift \parencite[][]{lass_what_1992}, and the Norman Conquest led to the replacement of a large number of Middle English words with French alternatives \parencite[][]{durkin_borrowed_2014}. 
This kind of competition and replacement is core to the study of borrowing and innovation in historical linguistics and sociolinguistics \parencite[cf.][]{mcmahon_understanding_1994,labov_principles_2011,mufwene_competition_2002,croft_explaining_2000}, to discussions of linguistic selection and drift \parencite[][]{baxter_modeling_2009,cuskley_internal_2014,sindi_culturomics_2016,newberry_detecting_2017,turney_natural_2019,pagel_dominant_2019},
S-shaped curves in language change \parencite{blythe_scurves_2012,ghanbarnejad_extracting_2014,stadler_momentum_2016,feltgen_frequency_2017}, and to studies of lexical growth and competition in big data computational linguistics \parencite[cf.][]{altmann_niche_2011,stewart_making_2018}.\footnote{Orthogonal to this is competition between entire languages or varieties \parencite{abrams_modelling_2003,castello_agentbased_2013,zhang_principles_2013,karjus_testing_2018}.}
These population-level approaches, which track competition between variants over potentially substantial time spans in populations of many individuals, are complemented by  the psycholinguistic literature studying competition between representations within individual brains \parencite[cf.][]{macwhinney_competition_1989,brouwer_can_2012,mickan_betweenlanguage_2020}. The choices of individual speakers are of course what constitute synchronic variation, which in turn may accumulate as changes observable on the level of the larger community consensus over time.

\begin{figure}[tb]
	\noindent
	\includegraphics[width=\columnwidth]{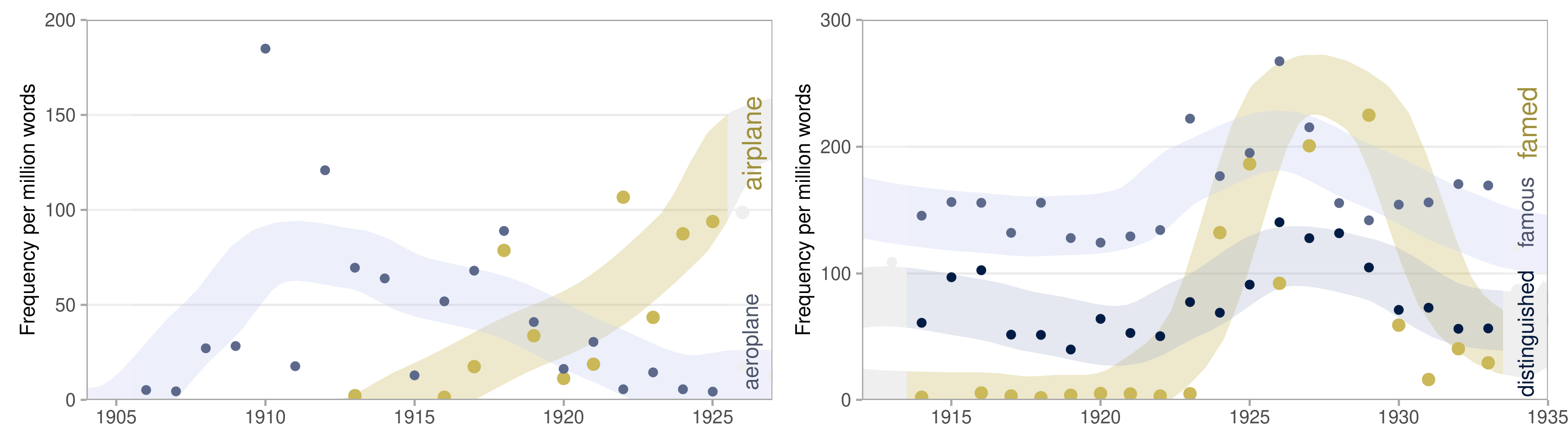}
	\caption{
		Example time series from the Corpus of Historical American English (COHA). Two decades after the invention of heavier-than-air powered aircraft, \textit{airplane} replaced the initial term \textit{aeroplane} (left side panel; the points are normalized yearly frequencies, with the lines representing smoothed averages for visual aid). 
		Around the same time, \textit{famed} appears to be increasing in usage. Yet it does not replace any semantically close words --- \textit{famed}, \textit{famous} and \textit{distinguished} all increase in tandem. Why do some successful new words replace their near-synonyms when their usage spreads, yet some do not, instead enriching their immediate semantic space?
	}\label{fig_firstexample}
\end{figure}

We make three contributions in this paper. First, we propose a quantitative model of competition between linguistic elements in large-scale diachronic corpus data. Then, we use this to demonstrate that competition dynamics are modulated by communicative needs of speakers. Finally, we argue that not all competition takes place between individual elements like words but rather collections of elements (topics of conversation).
As illustrated in Figure~\ref{fig_firstexample}, while some linguistic innovations lead to direct competition between synonymous variants (like \textit{aeroplane} and \textit{airplane}), potentially resulting in the eventual decline and replacement of all but one of the competing forms, many cases of innovation do not. Figure~\ref{fig_firstexample} gives as example the non-competition between \textit{famed} and {\em famous}, near-synonyms that increased in frequency in lock-step in the 1920s, at around the same time that \textit{airplane} was replacing \textit{aeroplane}.  
In short, there is variation in the presence or nature of competition --- some words like \textit{airplane} which enter a language or spread beyond niche usage compete with and replace a similar word, and may end up being replaced themselves in the future, whereas some words like \textit{famed} seem to exist companionably alongside other closely-related words. 

Our hypothesis is that communicative need affects the nature (specifically the ``directness") of the competition between individual words. We regard communicative need as a property of a topic of conversation, i.e. a subject consisting of related themes and ideas, encompassing a subset of co-occurring vocabulary.
When communicative need within a topic is constant --- its importance to a language community is not changing rapidly -- then any newly introduced words must compete with words with similar semantic functions that are already present in the language. This is the case for the topic of early aviation which \textit{airplane} belonged to in the first decades of the 20th century (cf. Figure~\ref{fig_adv} in Section~\ref{sec_methods_commneed}).
In contrast, where communicative need is on the rise --- a topic is increasing in importance for language users --- there is less need for competition between words, with multiple words able to co-exist and ride the wave of the users' communicative needs (like \textit{famed} and \textit{famous} do). 
We use computational methods (see Section~\ref{sec_methods}) to quantify the notions of topic, communicative need, and directness of competition. While our focus here is on the lexicon, we believe that given the general nature of our proposed approach to quantifying competition dynamics, it could be applied to other areas of language like syntax or phonology, provided a sufficient quantity of suitably annotated data is available.

Our hypothesis, as stated above, is informed by prior work arguing that the shape of the lexicons and grammars of natural languages reflect the communicative needs and preferences of users of a given language. This idea has a long tradition, going back to \textcites[: 26]{boas_mind_1911}[: 228]{sapir_language_1921}[: 2]{martinet_function_1952}. 
These needs are unlikely to be uniform across languages and time, or as \textcite{lupyan_why_2016} put it, ``aspects of language that promote its learning and effective use are likely to spread, but what is optimal for one environment may be suboptimal for another". 
This view is widely shared by authors discussing communicative needs as a possible driving force in language change \parencites[: 117]{givon_tenseaspectmodality_1982}[: 118]{arends_gradualist_1994}[: 74]{tomasello_cultural_1999}[: 37]{hopper_grammaticalization_2003}[: 286]{frajzyngier_explaining_2003}{vantrijp_selfassessing_2012}{mufwene_language_2013}{dor_instruction_2015}{kemp_semantic_2018}{winters_languages_2015}{winters_contextual_2018}{altmann_niche_2011}. 
For example, languages are known to vary in the number of colours they lexify and how elaborate their kinship vocabulary systems are, which has been argued to reflect differences in communicative needs of linguistic communities \parencite[][]{gibson_color_2017,zaslavsky_color_2019,kemp_kinship_2012}, and languages in warm climates are more likely to have a single word for both \textit{ice} and \textit{snow}, while these are lexified as individual words in colder climates \parencite{regier_languages_2016}.
Similar environment-driven effects have been shown to operate in number marking systems \parencite[][]{haspelmath_explaining_2017}, and proposed as a possible driver of colexification dynamics besides conceptual similarity \parencite{xu_conceptual_2020}.

Perhaps the most obvious locus where (semantic) communicative need can lead to change in any given language is a lexical gap --- a semantic subspace lacking an expression \parencites[][: 157]{trask_dictionary_1993}[][: 79]{blank_why_1999} or occupied by a word that has lost its expressive force \parencites[][: 201]{mcmahon_understanding_1994}[][]{tamariz_cultural_2014}. This gap may be filled with a suitable word or construction, either innovated within a language, or borrowed from another language, often from a socially more prestigious one \parencite[cf.][]{hernandez-campoy_handbook_2012,monaghan_cognitive_2019,calude_modelling_2017}.

In our case, it is this more local and transient sense of communicative need we seek to measure and explore, specific to a given language and a given culture in a given population at a particular time in its history.
This is in contrast to the broader sense of communicative need of languages being required to meet certain general criteria to be both learnable and useful as tools of communication \parencite[e.g.][]{zipf_human_1949,labov_building_1982,christiansen_language_2008,kirby_compression_2015,dingemanse_arbitrariness_2015,auer_hinskens_2005}. %

The motivation for our argument on competition between collections of linguistic elements stems from these more general communicative needs.
Given the pressure on languages to be learnable and efficient systems of communication, it would be reasonable to expect that the increase of complexity in one part of the lexicon (i.e. the entry of a new lexical item) would require a compensating simplification elsewhere. This could be either in the same lexical subspace, i.e. the new word replaces a semantically similar word --- or elsewhere, i.e. the incoming word is associated with a topic experiencing high communicative need, driving out words associated with other topics. In contrast to word-level competition, to our knowledge these dynamics have been given little attention in language change literature.
\textcite{karjus_quantifying_2020} demonstrate that individual word frequencies tend to follow the fluctuations of topics over time, as observable in population-level aggregate data such as corpora\footnote{\textcite{hofmann_predicting_2020} find a similar correlation between morphological family size and
topical dynamics.} (but presumably also in the differential salience of topics in the minds of the individual). For example, in times of war people talk about war-related things using more detailed vocabulary than they would otherwise; around major sports events like the Olympic Games various sports-related terms occur more frequently. These topical fluctuations can be taken to reflect the changing communicative needs of language users, reflecting the things that language users want and need to use language to communicate about.

We combine four distinct computational components in order to test our main hypothesis that communicative need affects lexical change and measure the directness of competition, ranging from word-level to topic-level.
The first step is to collect samples of words which we will refer to as ``targets" (Section~\ref{sec_targets}). This is the test set for the model, words which have increased considerably in usage frequency over some period of time, and possibly replaced some other words. 
Our focus on the lexicon is partially driven by technical challenges: words are by far the most straightforward to model using the lexico-statistical machinery we employ to infer meaning from data, given the current state of available tools and datasets.

Word similarity is operationalized by training a distributional semantics model where distances between all words can be measured (Section~\ref{sec_methods_comp}). Words similar to the targets will be referred to as ``(semantic) neighbors", referring to proximity in semantic space. This is a more suitable term than ``synonyms", as our unsupervised machine learning approach conflates various possible semantic relations --- such as synonymy, antonymy, hyperonymy, associativity --- into a single similarity metric. 

Previous research has pointed at the difficulties of capturing competitive dynamics and its effect on word growth \parencites[][: 155]{grieve_natural_2018}[][: 4368]{stewart_making_2018}. We propose an approach supported by machine learning to solve this (Section~\ref{sec_methods_comp})
We identify the locus of competition by summing up frequency changes in the ordered set of words similar to a target word, inferred from our model of semantics. In cases of competition between related words, the increase in frequency in the target word will be balanced by a decrease in frequency in a close semantic neighbor, whereas in more indirect inter-topic competition the frequency change will be balanced by decreases in distant, even unrelated, words.

As the final step, communicative need is inferred by proxy \parencite[as proposed in][]{karjus_quantifying_2020} using a simple information-theoretic topic model (Section~\ref{sec_methods_commneed}). Both this and our measure of word similarity rely on different operationalizations of word co-occurrence statistics --- we take care to ensure that these measures do not cause autocorrelation in the final explanatory model.
Our approach does involve a number of parameters and technical choices when it comes to training the machine learning models and operationalizing the corpus data. However, we find the results of our approach to be fairly robust within reasonable parametrizations.
We apply these techniques to corpora spanning 5 language varieties and 3 centuries. We find a small but significant effect repeating across all five datasets, supporting our hypothesis that communicative need modulates competition. Words that are introduced into language or disseminate beyond occasional niche usage are more likely to take over the semantic functions and cause a decrease in the usage of neighboring words, if communicative need in their topics remains stable. On the other had, if communicative need is elevated, they instead enrich the semantic space without causing their semantic neighbors to go out of use.

\FloatBarrier

\section{Methods and materials}\label{sec_methods}

\subsection{The corpora}

\begin{figure}[bt]
	\noindent
	\includegraphics[width=1\columnwidth]{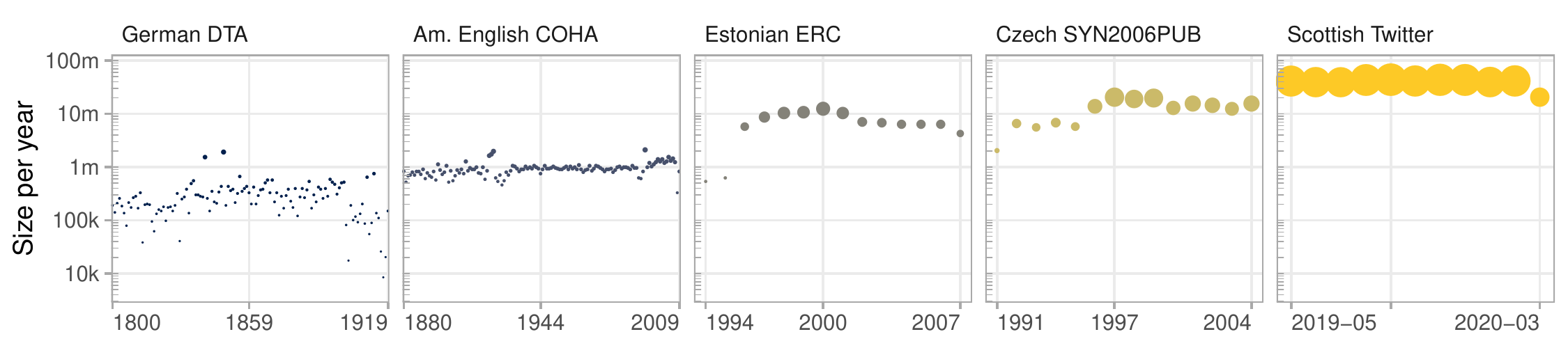}
	\caption{
		An illustration of the variability of the corpora as used in this study. The points reflect token counts by year (after filtering out stopwords; also note the $\log_{10}$ scale). The Twitter corpus of 315 days has over 3 times more data than 130 years of COHA put together (the Twitter panel shows monthly counts instead of yearly).
	}\label{fig_corpora}
\end{figure}

We test our hypothesis on data from five corpora, mostly selected by availability, but intended to cover a variety of corpus types, languages and time periods, as illustrated in Figure~\ref{fig_corpora}.
The Corpus of Historical American English \parencite[COHA;][]{davies_corpus_2010} spans the years 1810-2009 and includes 400 million words, balanced between four genres (newspapers, magazines, fiction and non-fiction). We use only data from 1880 onward as this part of the corpus is better balanced and more homogeneous.%
The Deutsche Textarchiv (DTA)\nocite{dta_deutsches_2019} %
is a corpus of German spanning 1600-1919, comparable in size and genre composition to the COHA. We use data from 1800 onward for the same reasons of genre balance and homogeneity as given for COHA. The Estonian Reference Corpus \parencite[ERC;][]{kaalep_estonian_2010} is a corpus of modern Estonian; we use the media and fiction parts between 1994-2007, most of the data consisting of daily newspapers. The SYN2006PUB \parencite{cermak_syn2006pub_2006} is a corpus of Czech newspapers between 1989-2004; we omit the first two years which have little data.

To diversify our test sets, we also mined Twitter for 315 days between 2019-2020 for all tweets posted in Scotland, and compiled it into a lemmatized corpus of 431 million words. Despite the idiosyncratic nature of Twitter communication, previous research has shown it to be a useful resource for studying language variation and change \parencite[cf.][]{grieve_mapping_2018,goel_social_2016,kershaw_modelling_2016}.
The timespan of this corpus is obviously much shorter than the traditional diachronic corpora, but it provides magnitudes more data per unit of time, metadata on the source of each utterance, and yields insight into how widely the predicted relationship between competition and communicative need holds.

COHA, DTA, ERC and SYN2006PUB were conveniently already tagged and lemmatized, and underwent similar preprocessing. 
Since we are interested primarily in the content word lexicon, we filtered out stopwords (function words, numbers, punctuation, etc.), using custom stopword lists and part-of-speech tags as available in the corpora themselves. We lowercased all texts and excluded proper nouns (using POS tags), as what they refer to can vary arbitrarily both diachronically and synchronically (e.g. a \textit{Bill} can be a president or a man on the street).
We noticed some proper nouns still seeped into the test sets due to tagging errors, but 
did not filter any of them out post-hoc.
For our Twitter corpus, we excluded duplicates and retweets, lowercased and lemmatized the texts \parencite[using spaCy;][]{honnibal_spacy_2017}, filtered out stopwords and @-tags (usernames, being essentially proper nouns) and further homogenized the data by removing the \# from hashtags (\textit{\#brexit} presumably means the same as \textit{Brexit}, while multi-word hashtags like \textit{\#borisjohnsonlies} retain the meaning with or without the \#).

\subsection{Target words and time series}\label{sec_targets}

Historical corpora are noisy population-level aggregate samples of utterances produced over time. Instead of attempting to model the dynamics of entire time series of all lexical items in a corpus --- inevitably mostly based on small noisy samples --- we opt to select a small set of examples of significant usage frequency change between well-defined time spans. This ensures that what we are modelling is not corpus compilation sampling noise. The large size of our corpora provides the luxury of collecting a sufficient number of such cases. This means precluding potentially interesting competition dynamics at low frequencies, but our unsupervised machine learning approach to meaning requires substantial amounts of data to remain reliable \parencite[cf.][]{wendlandt_factors_2018,dubossarsky_outta_2017}.

\begin{figure}[tb]
	\noindent
	\includegraphics[width=\columnwidth]{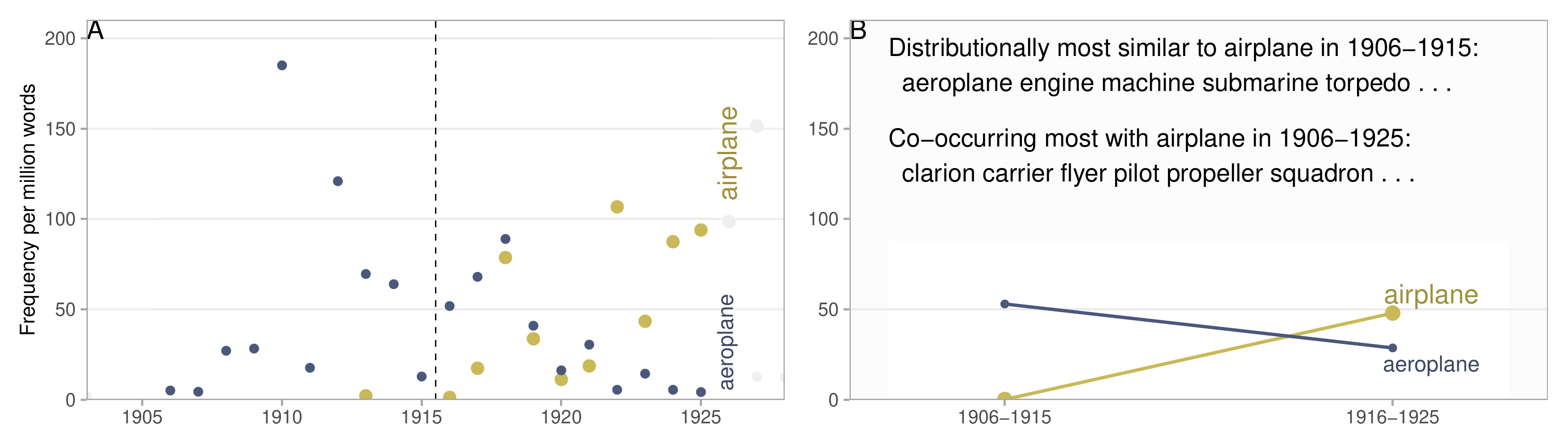}
	\caption{
		The \textit{airplane} data: yearly frequencies on the left and the binned competition model input on the right. Counts are aggregated into two 10-year bins. The dashed vertical line highlights the border between these 10-year spans. 
		The total increase of \textit{airplane} is almost matched by the decrease of its nearest semantic neighbor, \textit{aeroplane}, the remainder by decreases in \textit{engine} and \textit{machine}).
		The list of distributionally similar words on panel B include those that could be used instead of \textit{airplane} in a similar context (but not necessarily the same syntactic function or exact meaning; see Section~\ref{sec_methods_comp}). The list of co-occurring words include those that are used near \textit{airplane}, e.g. in a phrase like \textit{airplane pilot}, and form a basis for the model for inferring changes in communicative need (see Section~\ref{sec_methods_commneed}).
	}\label{fig_bins}
\end{figure}

We search each corpus for words that fit the criteria set out below. These criteria make reference to ``time spans'' and ``units of time'' which vary between the corpora, due to their different temporal resolution. We take the minimal time resolution within the corpora (days for Twitter, years for the others) to define the unit of time. The time span is a period over which a word's frequency is measured: each time span comprises multiple time units. 
For each potential target word $w$ at each unit of time in a corpus we evaluate the frequency change between (normalized) total frequency $f$ in the preceding time span ($t_1$) and the following one ($t_2$). For example, we use 10-year spans for COHA, so if the year considered by the search algorithm is 1916, then $t_1 = [1906,1915]$ and $t_2 = [1916,1925]$. In the \textit{airplane} example (cf. Figure~\ref{fig_bins}), the log (per millions words) frequency difference between $[1906,1915]$ and $[1916,1925]$ is $\ln(48/0.2) = 5.4$, which is the largest change for \textit{airplane} between any two 10-year spans in COHA.
We used 10-year spans in DTA; ERC and SYN2006PUB both span just over a decade, but contain much more data per year, so we used 5-year spans for those, and 30-day spans for the year-long Scottish Twitter corpus (see the Appendix for a longer discussion on the necessity of aggregating or ``binning" corpus data).

\pagebreak %

The selection criteria for a target word $w$ are as follows:
\begin{enumerate}
    \item Most importantly, the token frequency change of a potential target should be stark enough to cut through sampling noise ($\ln( f_{t_2} / f_{t_1} ) \geq 2$). 
    \item Its absolute frequency should be high enough for related statistics and distributional semantics inferences to be reliable ($f_{t_2} \geq 200 $). 
    \item In the Twitter corpus, where more metadata is available, we also require $w$ to be spread out across the user base (see Section~\ref{sec_controls} for details).
    \item $w$ should also be used throughout $t_2$ (in at least 80\% of units of time within the span of $t_2$)
    \item The frequency increase of $w$ should be consistent, the time series should not include outlying peaks (see Section~\ref{sec_controls} for technical details). 
\end{enumerate}
The last two criteria avoid cases where an apparent word frequency increase (simply based on comparing $t_1$ and $t_2$) stems from a word being a frequent term in some specialized corner of language or for a very short time period, while seeing little to no use in common language.
If multiple stretches along the time series of a word meet these criteria, we simply use the pair of time spans with the greatest frequency change between them; therefore each word only occurs once in the resulting dataset used for statistical analysis (Section~\ref{sec_results}).

This filtering procedure yields on average 270 target words per corpus 
(COHA: 240, DTA: 489, ERC: 274, SYN2006PUB: 257, Twitter: 97). Each target is associated with two time spans, between which the target word increased considerably in frequency, like the two decades of \textit{airplane} in Figure~\ref{fig_bins} (see Sections \ref{sec_results} and \ref{sec_methods_comp} for more examples of target words), and a number of lexicostatistical variables as described in the next sections. 
Further technical details on these parameters, and other implementation decisions, are discussed in the Appendix.

\FloatBarrier

\subsection{Modelling competition}\label{sec_methods_comp}

Our measure of competition derives from a simple notion about frequencies. After normalization, frequencies of words in a corpus sum to 1. Let us consider two possible sub-corpora from a historical corpus, e.g. for the 1990s and 2000s, each normalized separately. If a target word of interest increases between these periods, then it follows that some other word(s) must by definition decrease in frequency --- because in the end, everything sums to 1. In that sense, change in frequency always entails competition --- the increase of one word is matched by or ``equalized" by the sum of decrease(s) of some other word(s). Note that we use values multiplied by 1 million (per-million frequency) for more interpretable figures.

The distance to the target of the words whose decreases make up for the increase of the target can be taken as an indicator of the locus of competition. An increase that is directly compensated by an equivalent decrease in a semantically similar word (a near-synonym) indicates competition between two words that represent the same meaning. 
When no similar words decrease, and the increase in the target can only be matched by looking further away in the semantic space, then this indicates competition is less direct, and more likely to be between topics rather than within topics. We will refer to this distance --- where the frequency increase in a target is equalized by the cumulation of decreases in other words --- as the ``equalization range".

\begin{figure}[tb]
	\noindent
	\includegraphics[width=\columnwidth]{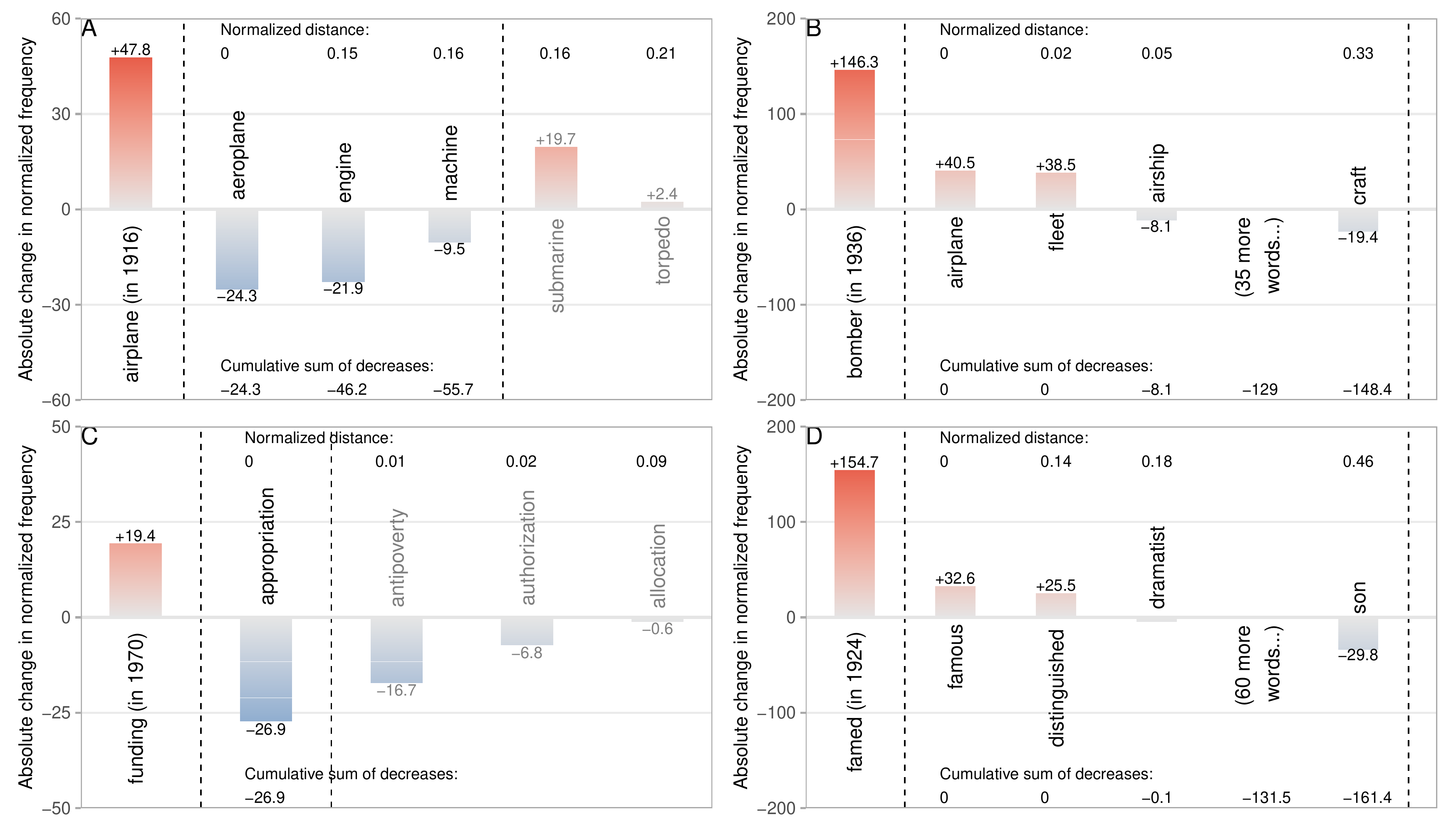}
	\caption{
		Examples of the competition model in action. The dashed lines indicate the equalization range. Panels A and C (the earlier \textit{airplane} example and \textit{funding} from the 1970s) are fairly clear cases of competition --- their increase is compensated by decrease in a near neighbor. For \textit{airplane} the equalization range is 0.16 (the normalized distance to \textit{machine}), as its increase of $+47.8$ per million words is matched by the sum of \textit{aeroplane}, \textit{engine} and \textit{machine}. The right side panels B and D illustrate lack of direct competition. \textit{bomber} came to prominence two decades after \textit{airplane}, when the topic of aerial warfare was too important to have only one word for all things with wings; note that \textit{bomber} emphatically does not compete with its nearest neighbor, \textit{airplane}, since both increase in frequency. For \textit{famed}, the equalization range is 0.46, and the decreasing words that compensate its increase are spread out between 60-odd words (note that at ranges this large, the measure simply indicates ``no direct competition" rather than reliably identifying actual competitors). 
	}\label{fig_comp}
\end{figure}

Importantly, this measure becomes less informative as its value increases. Large equalization range values should be interpreted as indicating that there are no direct (losing) competitors to be found, rather than considering the last equalizing word as a competitor. These are rather cases of what we suspect to be competition between topics --- which is of course much more indirect and hard to capture than competition between words with similar meaning. As the model searches for decreasing words further and further from the target in semantic space, the ones it finds may be quite unrelated to the target word (like \textit{son} in Figure~\ref{fig_comp}D). In contrast, at short equalization ranges, the competitors are usually clearly semantically related words (like \textit{aeroplane} in Figure~\ref{fig_comp}A).

Obviously, this approach requires some way of obtaining the similarity between all words in the corpus. 
This could be done using a dictionary \parencite[cf.][]{ramiro_algorithms_2018} or a lexical database such as a Wordnet \parencite[cf.][]{turney_natural_2019}, or using machine learning \parencite{xu_computational_2015,hamilton_diachronic_2016,rosenfeld_deep_2018,frermann_bayesian_2016}.
We opt for the latter approach, as this can be readily applied to any sufficiently large corpus, without the need for external language-specific resources. We use Latent Semantic Analysis \parencite[LSA; cf.][]{bullinaria_extracting_2007}, an application of Singular Value Decomposition. Like all distributional semantics models, it relies on word co-occurrence statistics. Words that are used together with the same words --- i.e. have similar distributions of co-occurrences across the lexicon --- end up with cosine-similar vectors in the resulting high-dimensional vector space. 
This acts as the computational approximation of a lexico-semantic space of a language, but with all semantic associations (cf. Section~\ref{sec_intro}) collapsed into a simplified proximity metric.

We measure the target equalization range using a normalized version of cosine distance. We observe that vector spaces trained on different corpora or even segments of the same corpus can have quite variable densities. Therefore it makes sense to normalize the distance, which we do by dividing by the distance value of the closest neighbor. Distance values of 0 in the results (cf. Figure~\ref{fig_results}) therefore refer to cases where the increase in frequency of a target word is completely matched by a compensating decrease in the nearest neighbor.

The LSA model is based on aligned co-occurrence data from the two time spans associated with a target word. The target is assigned a meaning vector using data from the second span where it is (thanks to the initial filtering) frequent enough, and the rest of the words in the lexicon are assigned meaning based on the first time span. There are two reasons for this. Since we require targets to undergo notable frequency change, most targets in the test sets have little to no presence before this increase, so it would not be possible to reliable infer their semantics. It is also not impossible that the increase of a target would change its immediate semantic landscape, forcing semantic change in related words \parencite[cf.][: 178]{mcmahon_understanding_1994}.
Our approach ensures the resulting semantic neighbors in the model are those that reside in the semantic space near the target just before its usage started increasing (see Figure~\ref{fig_comp}, and a more technical description in the Appendix).

The ease of operating with co-occurrence vectors and LSA in this manner is one reason to use this approach instead of a more recent model like word2vec combined with vector space alignment \parencite{mikolov_distributed_2013,hamilton_diachronic_2016,yao_dynamic_2018}. Our approach is analogous to the one described in \textcite{dubossarsky_timeout_2019,sagi_tracing_2011}, using common context words to model semantics over time. A context-sensitive model \parencite{devlin_bert_2019,hu_diachronic_2019} could potentially provide better meaning estimates, but would make comparing words between diachronic subcorpora less straightforward; this could be explored in future research.
Judging both by qualitative evaluation and testing against a gold standard test set \parencite{hill_simlex999_2015}, we found LSA to perform reasonably well despite the small size of time period subcorpora (distributional semantics models are usually trained on corpora of tens of billions of tokens, not mere tens of millions).

Our general approach is similar to \textcite{turney_natural_2019} who also investigate competition between words, but rely on the dictionary-like Wordnet data for determining similarity. Obviously inferring meaning using machine learning instead of using an expert-crafted lexicological resource has the downside of introducing additional noise. The upside is that since we infer meanings for words directly from respective time period sub-corpora, our approach does not require additional language-specific resources (such as a Wordnet), but also accounts for older and changed meanings (which a synchronic Wordnet does not). Furthermore, instead of modeling competition within predefined sets of synonyms (the ``synsets" of a Wordnet), our approach takes into account the entire lexicon with explicit similarity values, and allows us to account for indirect (topic-level) competition.

\FloatBarrier

\subsection{Modeling communicative need}\label{sec_methods_commneed}

Determining the communicative needs of the largely invisible speakers whose texts ended up in a historical corpus is by no means a trivial task. We estimate changes in communicative needs by assuming the following relatively simple model linking the observed corpus data and the presumed underlying process \parencite[see also][: 120]{kemp_semantic_2018}.

A diachronic corpus such as the COHA is essentially a large sample of utterances by numerous speakers (or more specifically, writers, in a written language corpus) expressing themselves across a variety of contexts and genres. 
If a topic of conversation is gaining importance for speakers, it would hopefully be reflected in the language, and therefore be observable as frequency changes in a representative corpus \parencite[assuming of course the apparent changes do not stem from sampling noise in an unbalanced corpus; cf.][]{pechenick_characterizing_2015}.
If the prevalence of a topic differs between two sub-corpora --- such as two decades --- then this can be taken to indicate differing communicative need within this topic.
If a topic is of socio-cultural importance to speakers --- the associated communicative need is elevated --- then it is reasonable to expect that speakers use the relevant vocabulary more, and use more detailed semantics for references in the discourse to successfully communicate more fine-grained distinctions, which may in turn result in the coining or borrowing of new words or repurposing old ones.
For example, the topic of \textit{bomber} (Figure~\ref{fig_comp}B), relating to aerial warfare, naturally became more prevalent during World War 2 --- which is reflected in widespread increases in frequency not only in \textit{bomber} itself but also in words it would co-occur (i.e., form a topic) with, such as \textit{squadron} or \textit{air force}, as well as the introduction of new ones such as \textit{blitz}.

\begin{figure}[tb]
	\noindent
	\includegraphics[width=\columnwidth]{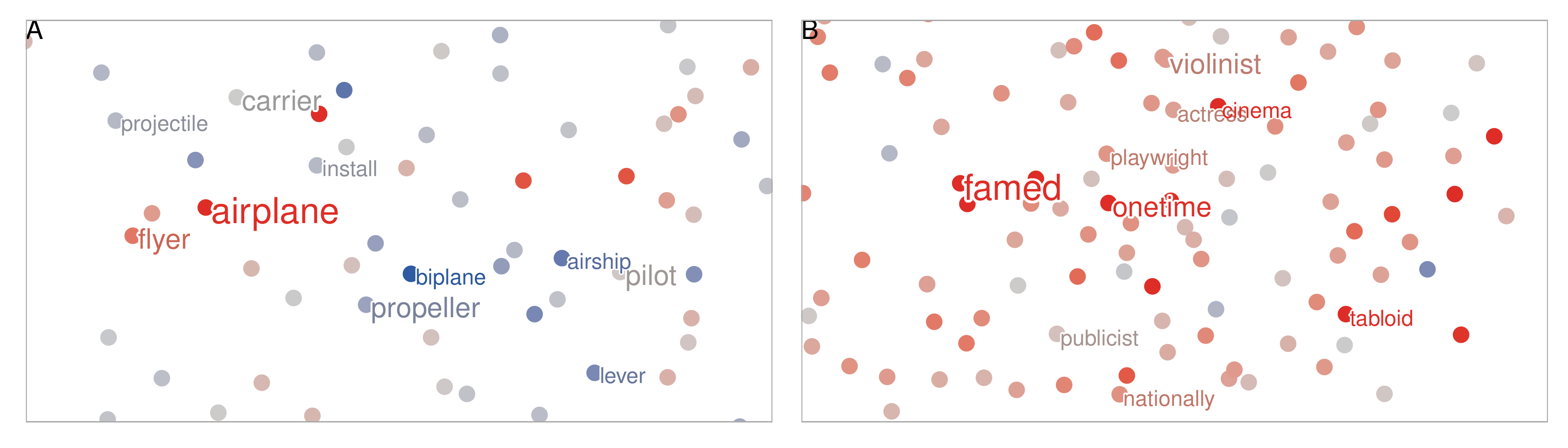}
	\caption{
		Topic landscapes for \textit{airplane} (in 1906-1925) and \textit{famed} (1914-1933). This is a dimension-reduced rough projection of the co-occurrence matrices of the subcorpora of these periods, with proximity between any two words approximately corresponding to not their semantic similarity but simply the extent which they occur together (more specifically, their PPMI association scores). A topic may be thought of as a group of words that are used together in similar contexts. Blue colors indicate words with decreasing frequency over the relevant time span, reds indicate increasing words. The advection value (weighted mean log topic change) for \textit{airplane} is close to neutral at 0.13, while it's strongly positive for \textit{famed} (0.82), reflected by being surrounded by plenty of red.
	}\label{fig_adv}
\end{figure}

We make use of the topical advection model from \textcite{karjus_quantifying_2020} to estimate changes in communicative need through quantifying the shifts in latent topics between time period sub-corpora. ``Advection" is a term borrowed from physics, referring to the transport of a substance by the bulk motion of a fluid --- the analogy being words swept along by prevalence fluctuations of associated topics. \textcite{karjus_quantifying_2020} show that this measure is a fair baseline predictor for word frequency changes --- it is possible to make a reasonable prediction about how much a word's frequency will change by looking at how well its related topic is doing. It is of course more informative for words that drift along with the flow of topics (such as \textit{famed} at the rise of cinema and celebrity culture in the interwar period; cf. Figure~\ref{fig_adv}) rather than those which compete with and are selected for (or against) by speakers, such as \textit{aeroplane}, which simply replaced a similar word with a similar spelling.

The topical advection model measures the change in topic frequencies between time periods (or sub-corpora more generally), not the prevalence of a topic at a given point in time.
We infer the ``topic" of each target word as a list of top $k$ context words which co-occur in the same context as the target (in a wider window of $\pm 10$ words), scored by their Positive Point-Wise Mutual Information (PPMI; see Appendix), as illustrated in Figure~\ref{fig_adv}.
Change in topic frequency is then measured as the weighted mean (log) frequency change of these topic words. 
Corpus data from both of the target's associated time spans $t_1$ and $t_2$ are concatenated, as this approach was shown in \textcite{karjus_quantifying_2020} to improve the model's performance.

At the heart of our approach are then essentially two non-overlapping lists of words: the list of (top 75 PPMI-scored) topic words --- and the list of semantic neighbors, ordered by similarity, spanning the entire lexicon (minus the topic words, to avoid autocorrelation).
Both lists are based on corpus co-occurrence statistics: topics consist of words occurring in the same context as the target, and semantic neighbors are essentially words which have similar context words. 
Sometimes a few of those may overlap: for example, besides \textit{aeroplane}, \textit{aircraft}, \textit{balloon} and \textit{propeller} all also have high similarity scores to \textit{airplane} --- but feature among its top topic words as well (cf. Figure~\ref{fig_bins}B), indicating co-occurrence in common contexts with \textit{airplane}.
It is crucial to avoid autocorrelation between the two measures --- which we do by filtering out such overlapping topic words (such as \textit{balloon}) from the list of neighbours when determining the equalization range.\footnote{The ease of decorrelating the measures this way is one reason to use a simple topic model based on discrete words here rather than something like LDA \parencite[][]{blei_latent_2003} which models topics as distributions; they were shown to perform comparably in \textcite[][]{karjus_quantifying_2020}. It is only feasible to do it this way around: the neighbours list spans the entire lexicon, while there are only 75 words in the topic list.
}
Leaving out topic words from the neighbours list unavoidably limits the descriptive power of the competition model: word(s) that sometimes occur in the same context with the target may also be among the ones that the target is actually in the process of replacing.

\subsection{Controlling for other lexico-statistical variables}\label{sec_controls}

We include a number of lexicostatistical measures as controls in the statistical model used to test the relationship between competition and communicative need. 
This is to exclude other possible explanations for variance in directness of competition, at least ones that can be inferred from a corpus (this unfortunately does not include possibly also relevant sociolinguistic variables).
Frequency change in the target (difference in per-million frequency values; cf. Figure~\ref{fig_comp}) is an important potential predictor: bigger increases could perhaps lead lower frequency neighbors to go out of use, or the opposite, bigger increases might require a larger equalization range. We also control for maximum (z-scored) peak value in the time series across the two time spans of each target (e.g. in COHA, yearly frequencies); the time point associated with the start of the increase in a target's time series (as a numeric value), and the length of the target word (long words might have different dynamics than short ones).

As for variables relating to the immediate semantic space, we control for minimum (Damerau-Levenshtein) edit distance of closest neighbors (is the target competing with a similarly spelled word?), cosine distance to nearest neighbor (does the target actually have close synonyms?), and the maximum percentage change among the nearest neighbors (does the target cause an extinction?). The last one differentiates cases of direct competition which lead to near-100\% decrease in a neighbor --- if the equalization range is short --- from changes which just lead to either a relatively small decrease in a high-frequency neighbor, or small decreases spread out between multiple neighbors. 

We also include a variable for leftover frequency mass (e.g. in Figure~\ref{fig_comp}, for \textit{funding} it would be $26.9-19.4=7.5$ units of per-million frequency, or 39\% of the $+19.4$ increase of \textit{funding}). If the decrease of the final equalizing neighbor is considerably larger than the increase in the target, then presumably either the model is not doing a good job capturing the semantics, or there is something more complex than just one-to-one competition going on (we also filter out targets where the leftover is actually larger than the increase value of the target). Additionally, in Twitter data, we control for the (median of daily) user to frequency ratio (see Section~\ref{sec_targets}).

For the Twitter corpus, we further make use of the available user metadata and, as mentioned in Section~\ref{sec_targets}, only consider targets which are reasonably widely used. Some words or hashtags may look very frequent at first, but a closer look often reveals (possibly automated) lone accounts or small groups that post the same or similar message hundreds of times a day e.g. to promote their views or products. This is of course not representative of common language use. We therefore excluded candidate targets with an account-to-frequency ratio\footnote{
I.e., the number of accounts who used a given term, divided by the total frequency of the term, yielding a value between $(0, 1]$. A ``1" means every occurrence is associated with a unique account; if 50 accounts each tweet a term twice (100 total), then it's 0.5; a single account tweeting a term 100 times yields 0.01. The filtering threshold uses the median of these daily values.} 
of $<0.75$, and also include this as a control variable in the statistical model for the Twitter dataset (see Section~\ref{sec_results}.
We did not make use of like and retweet counts, as they apply to entire tweets and not individual words --- although some averaged measure could potentially be considered in future research.

\FloatBarrier

\section{Results: communicative need predicts lexical competition dynamics}\label{sec_results}

\begin{figure}[b!]
	\noindent
	\includegraphics[width=\columnwidth]{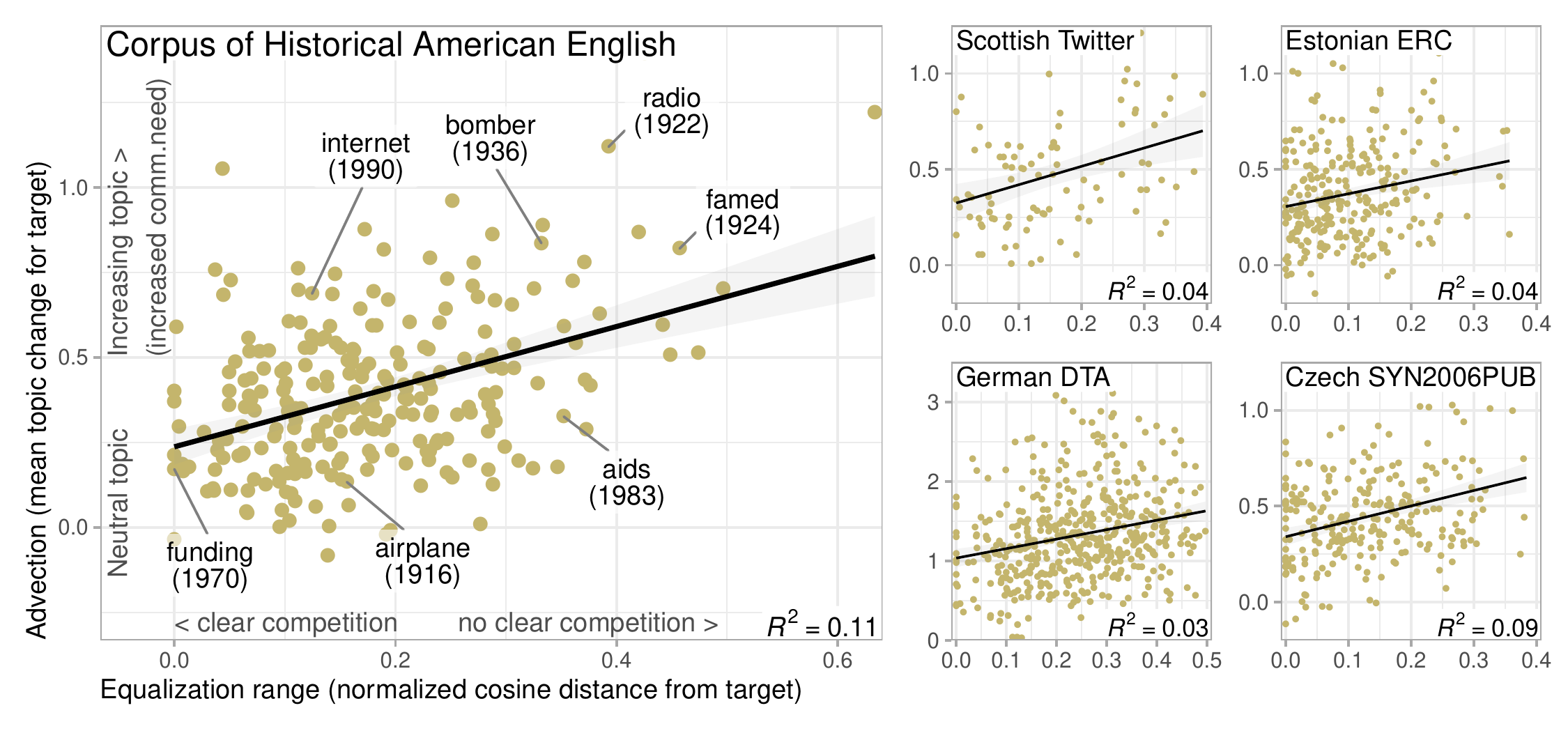}
	\caption{
		The extent of competition between words correlates with changes in communicative need, as operationalized by the topical advection model. The x-axis shows the equalization range --- where the frequency decreases of semantic neighbors match the increase in the target (cf. Section~\ref{sec_methods_comp}). Points on the left of the plot therefore represent words which compete and replace their immediate neighbor(s), whereas words on the right do not. The y-axis represents the mean topic change for the target word: advection values around 0 (lower on the plot) represent words in a topic which is relatively stable, whereas points higher up in the graph represent words whose topics are increasing in frequency in the corpus, which we assume reflects elevated communicative need around that topic.  
		There is a clear correlation between these two quantities (as reflected in points lying roughly on the plotted diagonal): words clustered in the bottom left corner are those in a stable topic and which compete and replace their immediate neighbor(s); words towards the top right corner increase in tandem with their neighbors, without a clear signal of immediate competition.
		The main panel shows this plot for the COHA corpus; the smaller side panels show that the effect of communicative need on competition dynamics persists with similar magnitude across data sets of target words based on corpora that differ in languages, time period and type of media.	 
	}
	\label{fig_results}
\end{figure}

Figure~\ref{fig_results} illustrates the results of applying the model to the target words extracted from our five corpora. We model the variables in a straightforward linear regression model, one for each data set. In all models, advection is a significant predictor ($p<0.05$) for the response variable of equalization range.
The $R^2$ values quoted in Figure~\ref{fig_results} refer to the amount of variance accounted for by the communicative need variable, on top of all the lexicostatistical controls described in Section~\ref{sec_controls} \parencite[adjusted $R^2$, based on comparing the full model to the reduced controls-only model, cf.][]{anderson-sprecher_model_1994}. As apparent in Figure~\ref{fig_results}, the model behaves comparably across the data sets, describing a moderate amount of variance (up to 11\%) in competition dynamics.  The German data set turns out to be somewhat of an outlier, with much higher absolute advection values, meaning that the topical composition of the corpus must fluctuate considerably over time. However, the correlation between advection and lexical replacement is still present. The full models have $R^2$ values between 0.17 and 0.25, as the lexicostatistical controls such as frequency change magnitude account for some additional variance.

If the usage frequency increase of a word --- or appearance in a language in the case of novel words --- does not coincide with a rising topic of conversation, then the word is more likely to take over the semantic functions of a similar word. 
For example, in American English, \textit{funding} encroached the semantic field of \textit{appropriation} in financial contexts in the 1970s, and \textit{boy scout} partially replaced the functions of \textit{cadet} in the 1910s after the founding of the namesake organization.  
In Estonian newspapers, it appears the term \textit{respublikaan} pretty much replaced \textit{koonderakondlane} from 2002 onward --- the former meaning `member of Res Publica', a center-right political party that became active in 2002, and the latter meaning `member of the Coalition Party', a center-right political party disbanded in 2002. This pair is of course not an example of synonymy, but reflects our model capturing terms used in very similar contexts to refer to similar political actors. 
In the Twitter corpus, \textit{movember}\footnote{
     An organization and annual event involving the growing of mustaches during November to raise awareness of men's health issues.
} starts trending towards the end of October 2019, replacing another charity-related term, \textit{greatscottishrun}; the rest of the increase is compensated by a slight decrease in the more frequent general term \textit{charity}.

In contrast, a word that increases in usage and belongs to a topic experiencing elevated communicative need is more likely to co-exist with synonymous or similar words. This would be the earlier \textit{famed} and \textit{bomber} examples, or \textit{radio} in the 1920s.
In the Twitter corpus, the term \textit{corona} occasionally pops up throughout the year referring to the beverage, but in the sense of the virus starts trending in January-February 2020 --- the pandemic of that year constituting a new high advection topic consisting of terms like \textit{virus}, \textit{spreading}, \textit{\#coronavirusupdate} but also the toilet paper emoji, all increasing in tandem with \textit{corona}.

\section{Discussion}

\subsection{Technical limitations and possible improvements}

We have shown that changes in the communicative needs of speakers contribute to lexical change and competition dynamics. However, we believe the real effect may well be larger than detected by our model. 
In addition to the peculiarities of written language corpora as discussed below (Section~\ref{sec_discussion_traces}), the models used here rely on statistical machine learning --- meaning, similarity and topics are all inferred from co-occurrence data. In other words, we rely on statistical approximations to communicative need and conventions, based on another proxy (corpora) to actual usage. Noise is unavoidable. The model is further weakened by the necessary purging of the semantic neighbors lists of often high similarity words to avoid autocorrelation with the topic model (cf. Section~\ref{sec_methods_commneed}). Yet we find the effect persists.

A reasonable worry would be that the small correlation between communicative need and competition dynamics we observe is a spurious one, an artifact of our statistical machinery, or some aspect of corpus composition. We do not have reason to believe so, based on carrying out simulations with randomized data on the competition model (see the Appendix), the advection model having undergone similar validation \parencite[cf.][]{karjus_quantifying_2020}, having controlled for a slew of other lexico-statistical variables, and having tested the model on a variety of different corpora.

There are several avenues of technical improvement that could be explored to build on the current contribution. 
These include using more sophisticated word embeddings (see Section~\ref{sec_methods_comp}), bigger corpora as they become available, and exploring the effects of different model parameterizations. 
In terms of corpora, investigating the role of communicative need in selection and competition in creole and new variety formation would be particularly interesting \parencite[cf.][]{baxter_modeling_2009,strimling_modeling_2015,winford_ecology_2017}.
Our essentially correlational results could be improved with causal analysis, and the methodology could potentially be extended to work with continuous time series \parencite[cf.][]{koplenig_datadriven_2017}.
The current competition measure identifies cases of direct competition, but becomes less informative as the equalization range increases. This calls for a method for more accurately inferring topic-level competition.
Connecting the competition model with tests for selection and drift could be explored \parencite[cf.][]{newberry_detecting_2017,karjus_challenges_2020,kauhanen_neutral_2017}.
Communicative need could perhaps be operationalized in ways that better approximate real world usage situations, possibly also by estimating diachronic developments via synchronic data \parencite[cf.][]{regier_languages_2016,karjus_spyglass_2015}.%

\subsection{Scarcity of direct competition}\label{sec_discussion_traces}%

We note that there are numerous examples among the target sets (cf. Section~\ref{sec_results}) where the equalization range consists of only a single neighbor. Yet examples of competition where the increase of a target word would lead to the complete disappearance of a neighboring one, at least within the timespan of a generation, are almost non-existent.  
It seems once a word has already entered conventional usage, it takes a while for it to completely disappear, even if it is on a clear downward path.
Even though \text{airplane} (beside just \text{plane}) is the preferred variant in American English, \textit{aeroplane} keeps popping up in the corpus throughout the 20th century, albeit at low frequencies, as does for example \textit{larboard} (the archaic nautical term for the left side of a ship) and \textit{cumbrous} (cumbersome). 
This echoes findings in previous research: while the entry of new linguistic material into language is often claimed to follow an S-shaped curve \parencite{blythe_scurves_2012}, extinction has been argued to follow a decelerated trajectory \parencite[][]{nini_application_2017}.

The unwillingness of words to die makes more sense if one considers the nature of written language corpora --- which may well include texts referring to historical events and objects, texts from more archaic varieties of a language (as British is to American English), and texts written (or edited) by older speakers for whom using older variants of modern terms comes naturally.
It has also been pointed out that the shape of the lexicon may not always reflect the current cultural interests and communicative needs of a community, with terms in semantic subspaces of waning relevance nevertheless surviving generations of speakers \parencite[][: 591]{malt_how_2013}.
Finally, there is also a further explanatory variable that we do not control for in our model: our approach to competition is based on usage frequencies, but there is also the possibility that a word losing out in competition might change meaning and continue to survive in another function (see the Appendix for details). Our simple model of semantics also treats each form as having a single (vector of) meaning, and competition may also resolve thought the loss or gain of semantic functions in polysemous words.

\subsection{Different kinds of competition}

Our findings point to language change being driven by yet another kind of competition in addition to those discussed in Section~\ref{sec_intro}. This is the competition between topics of conversation --- in turn presumably reflecting the events and state of the changing world. Word frequencies loosely follow topical fluctuations over time \parencite{karjus_quantifying_2020}, and our findings further illustrate that indeed many words that get introduced to language (or spread beyond previous niche usage) do not do so directly at the expense of older synonyms --- the nearest words that can be found decreasing in frequency are often semantically unrelated to the target. Instead, they follow the fluctuations of topics. 

Inevitably, when some topics of conversation increase in prevalence, others must diminish (there are only so many hours in a day). And in less relevant topics, semantic spaces will become sparse, as multiple words with slightly different shades of meaning become redundant, due to lowered communicative needs in the area. 
Historical and sociolinguistics often focuses on isolated examples of lexical replacement by borrowing or competition between language-internal variants. We believe competition between topics or semantic subspaces is something that deserves further investigation.
Furthermore, while grammatical complexity is widely studied and shown to correlate with population size and structure \parencite[][]{atkinson_speaker_2015,bentz_languages_2014,reali_simpler_2018}, linguistic topical complexity --- not just vocabulary size --- remains virtually unexplored.

\subsection{Using experimentation to further understanding of linguistic change}

Human language is a unique system seen nowhere else in nature. Understanding how and why it works requires understanding how it changes, change being one of the few absolutely universal properties of living languages. This in turn requires understanding both individual and population level dynamics. On the one hand, behaviour of linguistic communities is not necessarily indicative of the biases or choices of individual language learners and users, and different biases may lead to similar outcomes; on the other hand, constraints at the population level may arise from weak individual biases that may be hard to detect in isolation \parencite[][]{smith_eliminating_2010,smith_language_2017,kandler_inferring_2017}.

While the exact histories of the sociolinguistic environments where changes take place cannot be reconstructed, corpora, though imperfect lenses, provide a way to systematically observe wider changes in populations over time, like the growth and decline of elements of the lexicon.
The correlation we have observed calls for further investigation into the role of communicative need and fluctuations of topics in language change, also from the perspective of individual learning and communication biases. Unlike historical dynamics, this is something that can be studied in controlled experimental settings, either using natural \parencite[][]{lev-ari_experimental_2014} or artificial languages \parencite[cf.][]{kirby_cumulative_2008,winters_languages_2015,scott-phillips_language_2010}.

\section{Conclusions}

Previous research using experimental approaches and synchronic data has shown how languages adapt to the communicative needs of their speakers. We have shown how to model these processes and correspondences using data that reflects changes in language communities over longer time spans.
Our methods do not require language-specific resources other than a sufficiently large diachronic corpus, and produce comparable results across corpora of different languages, types, genres, and time spans.
In particular, we have described a language-agnostic approach to quantifying competition between elements of language, here on the example of lexical items. We found that these dynamics correlate with changes in communicative need, as operationalised by the topical advection model. 
In summary, we find support for the idea that languages keep changing in ways that are useful for their speakers. 
All other things being equal, multiple similar words can co-exist in a lexicon as long as the finer shades of meaning they provide are useful in discourse --- while new words will eventually replace old ones if a single word will do in the given semantic subspace.

\section*{Acknowledgments}

We thank Steven Piantadosi for a discussion that led to the operationalization of the competition model (cf. Section~\ref{sec_methods_comp}) and Jennifer Culbertson for useful questions and comments.
The first author of this research was supported by the scholarship program Kristjan Jaak, funded and managed by the Archimedes Foundation in collaboration with the Ministry of Education and Research of Estonia.

\section*{Data and code availability}

Some of the corpora are publicly available (see respective references below). The code to run the models described in this paper is available at \url{https://github.com/andreskarjus/competition-langchange}

\printbibliography

\FloatBarrier

\section*{Appendix}

\subsection{Using corpus data requires some form of aggregation}

The minimal time resolution in most of the corpora we used is one year. However, there is not enough data in most diachronic corpora per year for word embedding models (which we use to estimate word semantics) to work as intended. For COHA and DTA we used time spans or ``bins" of 10 years. ERC and SYN2006PUB both span just over a decade, but contain much more data per year, so we used 5-year spans for those. For the year-long Scottish Twitter corpus we used 30-day spans.
The limitation of comparing pairs of discrete time spans is of technical nature: the version of the topical advection model (Section~\ref{sec_methods_commneed}) that we use as a proxy to communicative need is not readily applicable to continuous time series.

Instead of simply using fixed calendric spans e.g. decades or months, we carry out binning for each word separately, depending on where a word starts increasing.
Although a choice like splitting a centuries-spanning corpus into 10-year spans or a year into 30-days spans might feel intuitive, all these choices really are quite arbitrary. Binning has indeed also been shown to affect statistical models based on corpus time series \parencite{karjus_challenges_2020}. While we remain reasonably confident in the results produced in this paper, values of parameters like this --- and the ones discussed in Section~\ref{sec_targets} and further below in the Appendix --- is something that should be critically evaluated in future research.

\subsection*{Parameters and alternative setups}

The complexity of our approach to lexical competition, necessitated by the complexity of the linguistic processes and the challenges of estimating these from diachronic data, entails a number of relatively arbitrary parameters and design choices. We describe the results of what we consider an intuitively reasonable set of choices, but further research could further explore the parameter space. We also explored slightly longer time spans in COHA where this is possible (20 years, yielding similar results), and flipping the model to quantify the ``losers" of competition instead: in the main text, we focus on words increasing in frequency, which provide a clearer case for competition, as discussed in Section~\ref{sec_discussion_traces}. The model works the other way around as well, with targets being words decreasing in frequency, above some chosen change threshold --- in the COHA data, we find a significant but even smaller correlation between equalization range and advection for words going out of usage.

\begin{figure}[htb]
	\noindent
	\centering
	\includegraphics[width=0.7\columnwidth]{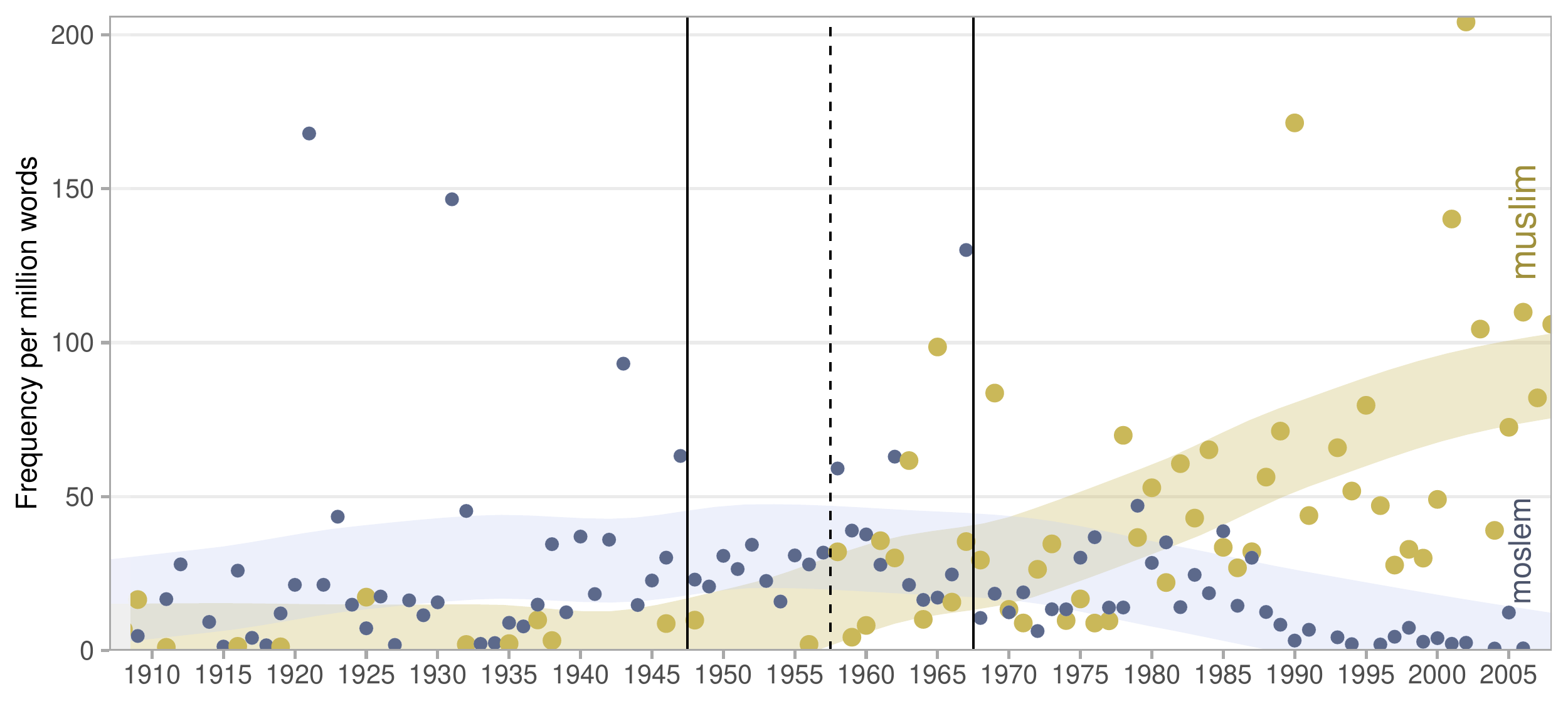}
	\caption{
		More example time series from COHA; points are normalized yearly frequencies, lines reflect smoothed averages. The black vertical lines indicate the period captured by our model as the largest (log) change between any 10-year spans in the time series of \textit{Muslim}. The outliers tend to coincide with related historical events. 
		This graph further illustrates the complexities of lexical change over longer periods. 
	}\label{fig_appendix_words}
\end{figure}

There is no doubt that some changes in language take more than our chosen time spans (e.g., 10~+~10 years).
There is no technical reason why we could not use longer time spans like 50 years, or compare a decade to another decade 100 years earlier --- but the worry is that multiple processes may well take place within longer periods, which our competition model (Section~\ref{sec_methods_comp}) is not tailored to handle.
Figure~\ref{fig_appendix_words} illustrates this, how very similar words can go through periods of co-existence and competition, and how competition can play out over longer time scales. \textit{Moslem} occurs at low frequencies in COHA throughout the 19th century, \textit{Muslim} appearing in the beginning of the 20th. These spelling variants then co-exist for half a century still at low frequencies (with a few exceptions like that of 1921, the year of the Malabar Rebellion), the former slightly more frequent than the latter, until \textit{Muslim} starts increasing --- but it still takes another 30 or so years until \textit{Moslem} starts decreasing.

\subsection*{PPMI and log change}

We employ two useful metrics in multiple stages of our model, namely Positive Pointwise Mutual Information (1) and log frequency change (2). Both are used in the advection model (see Section~\ref{sec_methods_commneed}), PPMI is used to detect multi-word units (see below), and log change is used to filter target words that have increased considerably between two time spans $t_1$ and $t_2$

\begin{flalign}
    {\rm PPMI}(\omega, c) := \max\left\{ \log_2 \frac{ P(\omega,c) }{P(\omega)P(c)}, 0 \right\} &&
\end{flalign} 
\begin{flalign}
    {\rm logChange}(\omega;t) := 
	\begin{cases}
	   0  & \text{if $f(\omega;t_1)=0$ and $f(\omega;t_2) = 0$ } \\
	   \ln[f(\omega;t_2)+s_{t_2}]-\ln[f(\omega;t_1)+s_{t_1}] & \text{otherwise}
    \end{cases} &&
\end{flalign}

$\omega$ is a word and $c$ is a context word it may co-occur with. $f(\omega;t)$ is the (in our case, normalized) frequency of $\omega$ in the corpus during time period $t$. We use Laplace smoothing offset to avoid $\ln(0)$, setting the values of $s$ to the equivalent of 1 occurrence in $t$ in after normalizing to per-million counts; $s$ is set to 0 if the frequency $f(\omega;t)>0$. If both frequencies are 0, then change is set to be 0.
While we refer to (2) as ``log change" \parencites[cf. also][]{altmann_niche_2011,petersen_statistical_2012}, it is also called log percent \parencites{tornqvist_how_1985,wetherell_log_1986} or logarithmic growth rate \parencite{casler_why_2015}.
Like absolute change ($f(\omega;t)-f(\omega;t-1)$), but unlike percentage change, log change is symmetric and additive. However, on the absolute scale, the biggest frequency changes are those of fluctuating already high-frequency words, while log change highlights sharp and therefore hopefully meaningful changes at lower frequency bands. This, in combination with chosen minimum increase threshold, results in target sets that mostly start out at 0 or low frequency (in COHA, the median across targets in $t_1$ is 1.9 per million words, or about 19 occurrences in an average decade of 10m words) and reach frequencies reflecting widespread usage by $t_2$ (median 30.1 pmw in COHA).

\FloatBarrier

\subsection*{Spelling smoothing and multi-word units}

We homogenize spelling by removing all punctuation (including hyphens) from lemmas in all corpora, and in COHA, concatenate the most common multi-word units, which we detect as follows.
The latter motivated by the fact that the spelling of compounds in English varies both diachronically and synchronically (e.g. \textit{long term}, \textit{long-term}, \textit{longterm}). This was done with the aim of improving our co-occurrence based measures of synonymy and topicality: in compounds (or phrases, collocations) such as \textit{social worker} or \textit{death row}, the words on their own often have different or at least more general meaning. Lexical innovations such as \textit{website} (also occurring as \textit{web site}) often go through multiple variants, making it harder to track their spread. In this contribution we focus on (homogenized) lemmas, leaving competition between low-frequency spelling variants like that for future research (one that would likely need larger corpora in terms of data per unit of time).

In multi-word unit detection, we only consider two-word units to keep things simple. As the first pass, the 200-year COHA is split into 20-year subcorpora and in each one, common multi-word units are determined using PPMI as the collocation metric (with a threshold of 7). We do not do this using the entire corpus at once, as collocation statistics may well change over time. The union of these 10 sets yields a total of 501 units, mostly compounds such as \textit{post office}, some phrases like \textit{absolutely necessary}, and a few proper nouns like \textit{Gulf War}. On the second pass, when parsing and cleaning the corpus for the analysis proper, these multi-word units are concatenated when encountered (e.g., to become \textit{postoffice}), and treated as single words in the subsequent frequency counts, semantic and topic models.
However, we find that this operation only marginally improves the power of the statistical model based on COHA data at the end of the pipeline (see Section~\ref{sec_results}).

\subsection*{Notes on Twitter}

Our Twitter corpus is slightly different from the others in that it covers communication on the platform by all users from a given geographical region, in a short time span. In contrast to written language corpora, this should reflect more ``natural", unedited, and relatively homogenous language use --- but then again Twitter is also only a narrow, situational slice of language. Its user base and demographics are not necessarily representative of the actual population, and the utterances expressed on Twitter are (hopefully) still only a subset of the daily utterances produced by its users.

Looking back at Figure~\ref{fig_results}, the number of targets in the Twitter dataset is relatively small. This is due to our stringent selection criteria for targets, one of which is consistent usage over the given time span. Looking at the data, many time series appear ``spiky" instead (see below for more on peak detection). A word or hashtag occurs rarely except for a day or two where its usage then skyrockets, often referring to some event, a piece of news, or a TV show. This of course which makes sense given the nature of Twitter, and we naturally do not expect to see considerable language change in the span of a year, but rather the topic-type competition discussed above.

\subsection*{Notes on control variables}

We control for edit distance between a target and its nearest neighbors to account for words which may potentially be competing with their spelling variants, such as \textit{airplane} (see Section~\ref{sec_controls}). This necessitates arbitrarily defining how near ``near" is, and we pick a range of 20 words. Maximum decrease percentage among neighbors also involves an implicit range parameter, but this is just set to be the same as the equalization range (Section~\ref{sec_methods_comp}), and as such varies from target to target.

Since we work with aggregated (binned) frequencies, we also account for differences in the time series within the aggregates by quantifying their maximum peak value --- seeing how some words increase steadily, while for some words, an apparent large increase in aggregated (e.g. decade) frequency stems from a single high-frequency peak on closer inspection.
Each frequency value in an examined time series (e.g. a 20-year span in COHA) is z-scored, using the mean and standard deviation of the rest of the series, i.e. excluding the value itself. We record the maximum of these z-scores, and during the target search phase (Section~\ref{sec_targets}) also exclude candidate series where the maximum is $>10$, indicating a series with a large outlying peak (10 standard deviations away from the mean). Such peaks can stem from sampling noise (a yearly subcorpus may for example include a book where some certain term is highly frequent) or real-world events which get a lot of coverage in the short term but do not affect the lexicon in the long term (as is very common in the Twitter corpus).

\subsection*{Details of the semantics model}

The LSA model is trained on a PPMI-weighted co-occurrence matrix based on corpus data from the first of the two time spans associated with each target word ($t_1$, cf. Section~\ref{sec_targets}), reflecting the semantic space of the language before the usage of the target started increasing.
We use a window of $\pm 2$ words \parencite[cf.][]{levy_improving_2015}, $k=100$ for LSA dimensionality, and a minimal occurrence threshold of 100 tokens.
Most targets in the test sets have little to no presence in $t_1$, which would hinder reliable semantic inference. We collect the lexicon-length co-occurrence vector for the target from the second time span ($t_2$) subcorpus where its usage is by definition widespread and frequent, align it to the lexicon of $t_1$, and then fit this into the $t_1$-trained LSA. 
This way, the resulting semantic neighbors are those that reside in the semantic space near the target just before its usage started increasing (cf. Figure~\ref{fig_comp}).

We remove words from these neighbors lists which do no occur widely in $t_1$ (threshold to occur at least in half of a time span). This filters out words that appear prevalent in a decade but only because of high frequency in a single year, often a single document like one book. This does not reflect widespread usage and is likely sampling noise.

\subsection*{Evaluating the approach using randomized data}

Since our approach to modeling competition relies heavily on machine learning, the natural worry is that the results may result from some unknown property or artifact of the underlying complex models \parencite[cf.][]{dubossarsky_outta_2017,wendlandt_factors_2018}, or be driven by some other lexicostatistical confounds such as frequency. We therefore include a number of plausible control variables in our statistical analysis (see Section~\ref{sec_controls}), set a frequency threshold to exclude low-frequency and therefore unreliable words (Section~\ref{sec_targets}), make sure our two co-occurrence-based measures do not overlap (Section~\ref{sec_methods_commneed}), but also evaluate the competition model by feeding it randomized data. 

The competition model relies on an ordered list of similarity-scored words, the closest of which could be considered near-synonyms of the target (see Section~\ref{sec_methods_comp}). We carry out a randomization test by giving each target word arbitrary semantic neighbors with arbitrary similarity scores (drawn from the distribution of actual similarities), but calculating the equalization range as usual (see Figure~\ref{fig_comp}). Under this randomization the closest neighbours of \textit{airplane} will be usually be unrelated words, for instance \textit{chocolate} or \textit{rabbit}, instead of \textit{aeroplane}. If advection (or proxy for communicative need) still correlated with the equalization range based on the assumption that \textit{airplanes} may be considered synonymous and competing with \textit{rabbits} then this would be reason for concern about the validity of the approach. However, we find that advection is a significant predictor ($p<0.05$) in less than 5\% of 1000 permutations of the model, as tested on the COHA dataset (i.e., as expected, given an $\alpha$ of 0.05), indicating that this is (hopefully) not the case.

\subsection*{Polysemy}

We operationalized two further control variable which we omitted from the main text, semantic change and polysemy. Both are somewhat complex and difficult to parametrize.
Polysemy (and homonymy) constitutes a commonly acknowledged weakness of type-based vector semantic models like LSA or word2vec, which collapse the possible multiple meanings of a word form into a single vector. We sought to estimate polysemy of target words and include it as a control variable, implementing the measure of ``dissemination" proposed by \textcite{stewart_making_2018}, which is used to model a proxy to polysemy using a linear regression model predicting the (log of the) number of words a word co-occurs with (in a window of $\pm2$ words) by its (log) frequency, with positive residuals indicating polysemy. We found a simple linear regression to yield an inadequate fit, improved by using a second-order polynomial. However, the initial results based on COHA data were not particularly intuitive, and as a control variable it did not turn up significant in the statistical model at the end of the pipeline, so we omit this from the analysis and the main text.

\subsection*{Semantic change}

The semantic change measure derives from our model of synonymy, which has diachronicity and context alignment already built in (see Section~\ref{sec_methods_comp}). Semantic change is simply a measure of the (inverse of) the similarity between the (context-aligned) vectors of words in the two time spans. Semantic change in targets cannot be estimated, as most are very low frequency in the first time span. Measuring change in nearest neighbors requires a similar range parameter as the edit distance variable, but only the semantic change of neighbors that occur frequently enough in both time spans can be estimated. 

Looking at the distribution of change values which indicate most words as slightly changing between decades, we suspect there is also likely some noise in the measure, possibly due to the relatively small size of the time period subcorpora (in machine learning terms anyway).
We carried out simulation experiments to probe a possible correlation between frequency difference and semantic similarity, as a proxy to frequency change possibly causing what would look like semantic change (which would be highly undesirable). We did this by taking the last decade of the COHA, making a copy, and randomly relabelling some occurrences of a sample of words from various frequency bands as \textit{word'} in the copy \parencite[similarly to the evaluation approach in][]{karjus_quantifying_2020}. This has the effect that nothing else except the frequency of the target words changes --- so if a measure of similarity between \textit{word} and \textit{word'} changes, then the given method of inferring semantic similarity (and change) must be frequency-biased, as it is in reality the exact same word. We measured both cosine similarity and the fact of \textit{word'} remaining the top closest neighbour of \textit{word}, for a range of simulated frequency differences between $-10\%$ to $-99.9\%$, and did this with a few different count-based vector semantics models --- LSA, but also full-length PPMI vectors, APSyn \parencite[][]{santus_testing_2016} and GloVe \parencite[][]{pennington_glove_2014}. We find that all those are to some extent frequency-biased \parencite[echoing findings on word2vec by][]{wendlandt_factors_2018}, at least given data of the size and composition of a COHA decade, but also that the results of LSA did remain relatively stable as long as the downsampled frequency did not fall below 100-200 (hence our choice of frequency thresholds for the context and target words).

Frameworks have been proposed to evaluate semantic change metrics \parencite[cf.][]{dubossarsky_timeout_2019,schlechtweg_wind_2019}, but given the complexities listed above and in order to keep the main text focused on the central question, we decided to omit modelling semantic change in this contribution.

\end{document}